\documentclass[twoside]{article}

\usepackage[accepted]{aistats2021}
\usepackage[round]{natbib}

\usepackage[utf8]{inputenc} 
\usepackage[T1]{fontenc}    
\usepackage{hyperref}       
\usepackage{url}            
\usepackage{booktabs}       
\usepackage{amsfonts}       
\usepackage{nicefrac}       
\usepackage{microtype}      

\usepackage{mathops}
\usepackage{cancel}
\usepackage{graphicx}
\usepackage{subfigure}
\usepackage{marvosym}
\usepackage{caption}
\usepackage{wrapfig}
\usepackage{fixmath}

\usepackage{color}

\usepackage{multirow}
\usepackage[T1]{fontenc}
\usepackage{lmodern}

\begin{document}

%
\runningtitle{Modeling Graph Node Correlations with Neighbor Mixture Models}

%

\twocolumn[

\aistatstitle{Modeling Graph Node Correlations with Neighbor Mixture Models}

\aistatsauthor{ Linfeng Liu \And Michael C. Hughes \And  Li-Ping Liu }

\aistatsaddress{ Tufts University \And  Tufts University \And Tufts University} ]

\begin{abstract}
We propose a new model, the Neighbor Mixture Model (NMM), for modeling node labels in a graph. This model aims to capture correlations between the labels of nodes in a local neighborhood. We carefully design the model so it could be an alternative to a Markov Random Field but with more affordable computations. In particular, drawing samples and evaluating marginal probabilities of single labels can be done in linear time. To scale computations to large graphs, we devise a variational approximation without introducing extra parameters. We further use graph neural networks (GNNs) to parameterize the NMM, which reduces the number of learnable parameters while allowing expressive representation learning. The proposed model can be either fit directly to large observed graphs or used to enable scalable inference that preserves correlations for other distributions such as deep generative graph models. Across a diverse set of node classification, image denoising, and link prediction tasks, we show our proposed NMM advances the state-of-the-art in modeling real-world labeled graphs.
\end{abstract}

\section{Introduction}
Many real-world applications have data organized in a graph. One important task in the analysis of graph-structured data is to model the discrete labels of nodes, which are either observed or hidden. Probabilistic graphical models \citep{wainwright2008} describe dependencies between nodes in an elegant and extensible way, useful in practice for predicting node labels or edges in ways that account for uncertainty. 

Among graphical models, a Markov Random Field (MRF) \citep{kindermann1980} is often used to describe a distribution of node labels on an undirected graph, capturing dependencies between nodes via unnormalized potential functions. One well-known variant of the MRF is the Conditional Random Field (CRF) \citep{lafferty2001}, which explicitly includes observed features \citep{murphy2012} in potential functions. While an MRF has a convenient model definition, running inference for an MRF is a hard computational problem. For example, it is hard to evaluate the marginal probability for even a single node's label; it is also hard to sample a single node's label from its marginal distribution. These computations often require the inference over the entire graph, whose runtime scales exponentially with the number of nodes for the kinds of graphs commonly used to model rich dependencies in applications \citep{wainwright2008}. To scale up, practitioners appeal to approximations such as \textit{mean-field} variational inference \citep{blei2017} that make strict simplifying assumptions.
There is a need for scalable methods that adequately account for correlations between labels.

Recently, graph neural networks (GNN) \citep{wu2019} have made remarkable progress in learning representations from graph data. These models take in node-specific features and then propagate messages along graph edges through several hidden layers \citep{kipf2017, hamilton2017, velickovic2018, xu2019}. In the last layer, each node gets a vector representation that encodes information about the local graph structure and its observed features.
While GNNs focus on learning informative node representations useful for label or link prediction, they typically assume labels are \emph{independent} given all observed node features.

A natural idea is to combine graph neural networks with a probabilistic model to learn an expressive model for node labels given node features. The recent Graph Markov Neural Network (GMNN) \citep{qu2019} combines a CRF model with a graph neural network and shows the obvious benefit of modeling node labels in a probabilistic manner. However, due to the difficulty of MRF inference, GMNN requires substantial approximations such as pseudo-likelihood maximization  \citep{besag1975} and mean-field variational inference, the latter of which depends on an inference neural network that is not easy to train.
In this work, we aim to devise an alternative probabilistic model for graph-structured data that can utilize GNNs for effective inference while overcoming these difficulties in inference.

We propose the Neighbor Mixture Model (NMM), a new probabilistic model that captures correlations in the distribution of node labels through an attention mechanism \citep{velickovic2018}. This new model assumes every node attends one of its neighbors in the graph (including itself) to get the latent parameter that determines its label distribution. This parameter sharing induces correlation between nodes that share an edge. Our NMM \textit{model} itself, a joint distribution over the labels of nodes, can be evaluated or sampled without any specific node ordering. NMM parameters can be efficiently parameterized by a GNN or another task-relevant neural network. The NMM thus leverages the synergistic strengths of its two key components: neural networks for extracting useful representations from the input features and probabilistic graphical models for capturing correlations between nodes. 

To scale to large graphs, we contribute an efficient variational inference method for the NMM, with a tractable lower bound on the likelihood of node labels in a training set which can be efficiently optimized to estimate parameters. We empirically show our proposed variational lower bound on this likelihood to be tight and thus capable of capturing useful correlations between labels, while not requiring any additional parameters beyond the original model.

Beside modeling observed node labels directly, the NMM can also serve as an amortized inference method to efficiently approximate other models (e.g. MRFs or deep generative models of graphs). Previous methods of amortized variational inference \citep{kingmaVAE2014, shu2018amortized, zhang2018advances} mostly reuse parameters across independent instances. GNNs are recently used to do inference on small graphs \citep{yoon2018inference, satorras2020neural}, but training these GNNs requires ``ground-truth'' results from another inference method. Unlike previous methods, an NMM can reuse parameters for \textit{dependent variables} on \textit{large} graphs.
We show how to train an NMM to minimize its KL-divergence to the target distribution, without requiring any ground-truth inference results or extra parameters that scale with the size of the graph.

We evaluate our proposed NMM on node label classification, prediction of joint label configurations for connected node pairs, image denoising, and link prediction. When directly fit to observed labels in a graph, NMM improves probabilistic predictions over state-of-the-art baselines. When used to approximate complex distributions such as deep generative models for graphs~\citep{mehta2019} where exact inference is difficult, our flexible NMM captures correlations that commonly-used mean-field independence assumptions cannot, while maintaining scalability.

\section{Background and Notation}
Here we establish the learning problem.
Let $G=(V, E)$ denote an undirected or directed graph with $N$ vertices, where $V = \{1, \ldots, N\}$ is the vertex set and $E$ is the edge set. Each node $i \in V$ is associated with a tuple $(\bx_i, y_i)$, where $\bx_i \in \mathbb{R}^F$ is a real-valued feature vector (observed for all nodes) and $y_i \in \{1, 2, \ldots C\}$ is a discrete label. We stack all feature vectors into matrix $\bX$ of size $N \times F$, and stack all labels in a column vector $\by$ with length $N$. Our goal is to model the distribution $p(\by | \bX)$, in a way that captures dependencies between labels $y_i, y_j$ that are ``neighbors'' in the graph. 

Let $n(i)$ denote the \emph{neighborhood} of node $i$. The definition of the neighborhood, which determines the range of correlation, is a model choice. In this work, we focus on the neighborhood containing the first order neighbors and the node itself: $n(i) = \{ j : (i,j) \in E \}\cup \{i\}$. Throughout this text, let $D$ denote the \emph{maximum degree} across all nodes in the graph, and let $\Delta^{k}$ denote the space of probability vectors with length $k$.

\section{Method}
\subsection{The Neighbor Mixture Model}
\label{sec:model}

The NMM is a generative model for labels $\by$ given node features $\bX$ that requires two key parameters: $\balpha = \{\alpha_i \in \mathbb{R}_{+}^C: i \in V\}$ and $\bL = \{\bL_i \in \Delta^{n(i)}: i \in V\}$. Each $\balpha_i$ is a label concentration vector for node $i$. Each $\bL_i$ is a probability vector over neighbors of node $i$. These two parameters are computed from a neural network, $(\balpha, \bL) = f(\bX)$, which we will discuss later in Sec.~\ref{sec:gnn_parameterization}. Given these parameters, the model formally defines $p(\by | \bX) = p(\by | (\balpha, \bL) = f(\bX) )$ by 
\begin{align}
    \bz_j &\sim \mathrm{Dirichlet}(\balpha_j), &\bz_j \in \Delta^C, ~~j=1, \ldots, N, \nonumber \\
    c_i &\sim \mathrm{Categorical}(\bL_i),     &c_i \in n(i), ~~i=1, \ldots, N, \nonumber \\
    y_i &\sim \mathrm{Categorical}(\bz_{c_i}), &y_i \in \{1, 2, \ldots, C\}.\quad\quad~
    \label{eq-general-dist-detailed}
\end{align}
The NMM captures correlations between labels $y_i, y_j$ by sharing a distribution parameter, vector $\bz_j$, between neighboring nodes. 
Vector $\bz_j$ provides the probabilities for a categorical distribution over $C$ possible node labels.
Each node has its own $\bz_j$, generated a priori from a Dirichlet distribution.
To generate its label, each node $i$ then selects one node $c_i$ from its neighborhood $n(i)$ and then ``borrows'' the vector $\bz_{c_i}$ from this chosen neighbor to sample its node label $y_i$ given this probability vector. This formulation with explicit $c_i$-s is easier for inference later, while the marginalization of all $c_i$-s below gives a clearer understanding of how this model induces correlations between neighbors $y_i, y_j$.

\parhead{Marginalizing neighbor indicators:}
After integrating $c_i$-s away, our model for $y_i$ given $\bz$ is:
\begin{align}
    y_i \sim \mathrm{Categorical} (\bu_i),  \quad \bu_i = \textstyle \sum_{j \in n(i)} L_{ij} \bz_j.
    \label{eq-dist-mixture}
\end{align}
In this equivalent formulation, we view vector $\bL_i$ as node $i$'s attention weights over its neighbors. Each node computes label probability vector $\bu_i$ as a weighted sum of probability vectors $\bz_j$ in its neighborhood. Each $\bz_j$ has a mixing weight $L_{ij}$. Two nodes with overlapping neighborhoods are tied together by sharing some mixture components $\bz_j$. The joint density of $\by$ given $\bZ=(\bz_j)_{j=1:N}$ is decomposable: $p(\by | \bZ) = \prod_{i} p(y_i | \bz_{n(i)})$. 
If we further marginalize out $\bZ$, labels $y_i$ and $y_j$ are correlated if $n(i) \cap n(j) \neq \emptyset$. While we focus on positive correlations, our model can be modified to model negative correlations (e.g. using a transform of $\bz_{c_i}$).

\parhead{Sampling node labels:}
We can draw samples of $\by$ by following the model definition in \eqref{eq-general-dist-detailed}: first draw samples of $\bz$ and samples of $\bc$ from their respective priors, then use these values to draw samples of $\by$. We can also leverage \eqref{eq-dist-mixture} to sample $\by$ without sampling $\bc$.

By \eqref{eq-dist-mixture}, drawing samples for a single node $i$ from its marginal $p(y_i)$ only requires a few vectors $\bz_{n(i)}$ in its neighborhood, so it is \emph{efficient}.  In contrast, drawing an exact marginal sample for a single node from an MRF usually requires inference over the entire graph. 

\subsection{Exact computation of marginal probability for small node sets}
\label{sec:exact_marginal}
Consider observing labels for a \emph{subset} of nodes $\tau \subseteq V$. Given known parameters $\balpha, \bL$, we wish to compute the marginal probability of the subset's labels $\by_\tau$. We can do this using the sum rule:

\vspace{-6mm}
\begin{align}
p(\by_\tau | \balpha, \bL) = \textstyle \sum_{\bc_{\tau}} p(\by_\tau, \bc_{\tau} | \balpha, \bL)
\label{eq-exact-py}
\end{align}
We can further write each term in that sum as marginalizing away all $\bz_j$ variables related to any neighbor $j$ of any node in the set $\tau$. 
Let $n(\tau) = \cup_{ i \in \tau} n(i)$ denote all such unique neighbor nodes. Then the probability of interest becomes:
\begin{multline}
p(\by_\tau, \bc_{\tau} | \balpha, \bL) =
	\int_{\bz_{n(\tau)}}
		\hspace{-1em} p(\bz_{n(\tau)} | \balpha )
		p(\bc_{\tau} | \bL)
		\prod_{i\in \tau} p(y_i | c_i, \bz_{n(i)}) 
\\ = \prod_{i \in \tau} L_{i, c_i} \cdot 
    \prod_{j \in \bc_\tau} \frac{\mathrm{B}(\balpha_j + \bs_j(\by_\tau, \bc_\tau) )}{\mathrm{B}(\balpha_j)},
    \label{eq-joint-y-c}
\end{multline}
where the integral simplifies due to Dirichlet-Categorical conjugacy (see the appendix for derivation). We count the number of nodes using neighbor $j$ of each class with vector $\bs_j(\by_\tau, \bc_\tau) = \sum_{i\in \tau: c_i = j} \mathrm{onehot}(y_i)$.
 $\mathrm{B}(\balpha) = \frac{\Gamma(\alpha_1) \Gamma(\alpha_2) \ldots \Gamma(\alpha_C)}{\Gamma(\sum_{c=1}^C \alpha_{c})}$ denotes the multivariate Beta function.

The computation of $p(\by_\tau, \bc_\tau | \balpha, \bL)$ in \eqref{eq-joint-y-c} takes time $O(C|\tau|)$. Thus, we can compute $p(\by_{\tau} | \balpha, \bL)$ in \eqref{eq-exact-py} by summing over all the $D^{|\tau|}$ possible configurations of $\bc_{\tau}$. When $\tau$ is a small set, then the overall computation is manageable (e.g. for one node it is $O(CD)$). In contrast, for an MRF the marginal of even a single node requires the inference over the entire graph, which is often much more expensive.

\subsection{Scalable approximation of marginal for large node sets}
\label{sec:scalable_marginal}

For large node sets  $\tau$, we need to appeal to approximate inference, as the exact computation of the marginal likelihood becomes infeasible. We use variational inference~\citep{wainwright2008,blei2017} and derive a variational lower bound $ \calL(\balpha, \bL) \le \log p(\by_\tau | \balpha, \bL)$:
\begin{align} 
\calL(\balpha, \bL) := \E{q}{\log \frac{p(\by_\tau, \bc_\tau | \balpha, \bL)}{q(\bc_\tau | \by_\tau, \balpha, \bL)}}.
\label{eq-learn-obj}
\end{align}
Note that the bound becomes an equality when $q(\bc_\tau | \by_\tau, \balpha, \bL)$ is the true posterior $p(\bc_\tau | \by_\tau, \balpha, \bL)$, but this is intractable. The key technical challenge is choosing an approximate posterior distribution $q(\bc_\tau | \by_\tau, \balpha, \bL)$ close to $p(\bc_\tau | \by_\tau, \balpha, \bL)$ that makes the bound tight yet has more affordable computation.

To address this challenge, we define the distribution $q(\bc_\tau|\by_\tau, \balpha, \bL)$ as a directed graphical model with conditional probabilities derived from the true joint $p(\bc_\tau, \by_\tau | \balpha, \bL)$.
Given an ordering $\pi$ of nodes in $\tau$, we define $q$ as:
\begin{align}
q(\bc_\tau | \by_\tau, \balpha, \bL) = \textstyle \prod_{i \in \tau} p(c_i | y_i, \bc_{<i}, \by_{<i}, \balpha, \bL),
\label{eq-qcy-form}
\end{align}
where we denote the parents of node $i$ within $\pi$ as $(<i) = \{i' \in \tau: \pi(i') < \pi(i)\}$.

The conditional $p(c_i | y_i, \bc_{<i}, \by_{<i}, \balpha, \bL)$ is calculated exactly by computing the joint for each of the  possible $c_i$ values in $n(i)$ and then normalizing.
Recall that neighborhoods have bounded size ($|n(i)| \leq D$), so this is affordable.
Each needed joint $p(c_i, y_i, \bc_{<i}, \by_{<i} | \balpha, \bL)$ can be efficiently computed from \eqref{eq-joint-y-c} using the subset $\tau = \{i\} \cup (<i)$.
Thus, our chosen $q$ is easy to evaluate and sample from.

Our ultimate lower bound objective \eqref{eq-learn-obj} requires an expectation with respect to $q$. 
We can estimate this expectation using $T$ Monte Carlo samples $\{ \bc_{\tau}^{(t)} \}_{t=1}^T$ from $q(\bc_\tau | \by_\tau, \balpha, \bL)$. 
Each sample can use a different ordering of nodes $\pi^{(t)}$, sampled from a uniform distribution over permutations. 
Then the objective can be viewed as an estimate of the average of lower bounds derived from all possible node orders.

\parhead{Explanation of the chosen $q$ distribution:}
The parameterization of $q(\bc|\by, \balpha, \bL)$ in \eqref{eq-qcy-form}  reuses the same $\balpha, \bL$ parameters as the NMM model, introducing \emph{no} extra parameters. Though this choice may not be the most flexible distribution possible, it has several advantages.
First, it can capture correlations between neighboring nodes (we do not assume each $c_i$ is independent, as typical mean field approximations would). Second, it reduces the number of variables that need to be estimated in the later optimization of the variational lower bound. Otherwise, we may need another neural network to parameterize $q$. Training two neural networks jointly can lead to severe solution quality issues such as posterior collapse  \citep{lucas2019}. Therefore, our chosen $q$ makes training our method far easier and more reliable than alternatives.

\subsection{Parameterization using Neural Networks}
\label{sec:gnn_parameterization}

We wish to achieve a parameterization of our NMM that is both \emph{scalable} and \emph{informed} by observed node features $\bX$.
As mentioned earlier, we use a neural network $f$ to determine the parameters: $(\balpha, \bL) = f(\bX; \theta)$. 
Here symbol $\theta$ denotes all network parameters; we assume $\theta$ has fixed size that does not depend on graph size (unlike $\balpha, \bL$).
By leveraging the strong representational power of neural networks, this construction can compute parameters informed by node features $\bX$.

When the model is defined on a graph, we use a GNN as the backbone of $f(\cdot; \theta)$. A GNN takes node features and propagates messages between graph nodes to produce a vector representation for each node.
For our NMM, let the GNN produce vectors of size $C + H$, where $C$ is the number of possible labels and $H$ is a free hyperparameter.
Denote the first $C$ entries of this output vector as $\bu_i \in \mathbb{R}^{C}$, the remainder as $\bv_i \in \mathbb{R}^H$.

We deterministically transform these GNN outputs $\bu, \bv$ to produce our parameters $\balpha, \bL$:
\begin{align}
\label{eq-GNN-construction-of-alpha-L}
&\{ (\bu_i, \bv_i): i \in V \}  \gets \mathrm{GNN}(\bX; \theta), 
\quad \\
&\balpha_i \gets \sigma(\bu_i),
L_{ij} \gets \textstyle \mathrm{softmax}\left( \omega^2 \frac{\bv_i^\top \bv_{j}}{ \|\bv_i\| \|\bv_{j}\|} + \gamma \delta(i=j)\right). 
\nonumber
\end{align}
 To construct $\balpha_i \in \mathbb{R}^C_+$, we cast the embedding vector $\bu_i$ of size $C$ to positive values using activation function $\sigma(\cdot)$. Recall from \eqref{eq-general-dist-detailed} that positive vector $\balpha_i$ determines the label marginal associated with node $i$.

To construct $\bL_i \in \Delta^{n(i)}$, the probability vector which determines correlations with neighbors, we use a self-attention transformation of the embedding $v_i$, where parameter $\gamma > 0$ increases the probability that $i$ attends to itself.
Scalar $\omega^2$ controls the sharpness of the distribution.
The softmax function is taken over all indices in the current neighborhood $n(i)$.

For our GNN parameterization, the learnable model parameters $\theta$ include all GNN weights, $\omega^2$, and $\gamma$. 
For special graphs such as grids for modeling image pixels, we can also use CNNs as the backbone.

\subsection{Parameter learning}
\label{sec:parameter_learning}

Given a graph with known labels $\by_{\tau}$ for a large subset $\tau$, we could estimate parameters $\balpha, \bL$ by maximizing $\calL(\balpha, \bL)$ in \eqref{eq-learn-obj}, which is a lower bound of the NMM's label marginal likelihood $\log p(\by_{\tau} | \balpha, \bL)$.
With the new scalable parameterization, we maximize over $\theta$ directly, and our objective is a lower bound of $\log p(\by_{\tau} | \bX)$:
\begin{align}
\hat{\theta} \gets
	\arg \max_{\theta} ~\calL(\balpha, \bL), \quad (\balpha, \bL) = f(\bX; \theta).
	\label{eq-overall-obj}
\end{align}
The gradient of $\calL(\balpha, \bL)$ with respect to $\balpha, \bL$ requires Monte Carlo estimation of gradients of expectations over discrete random variables (since our indicators $\bc_{\tau}$ will be sampled from $q$ given parameters $\balpha, \bL$). We estimate these gradients using the well-known REINFORCE estimator~\citep{williams1992}, also known as the score function trick. The optimization procedure converges well, but we could also explore improved estimators such as Rebar and RELAX \citep{tucker2017, grathwohl2018} in future work. 

\subsection{Node classification with the NMM}
 In graph node classification tasks, a subset of nodes $\tau$ have observed labels, and we need to predict the labels of the remaining nodes $\kappa \subset V$. In training, we learn an NMM defined by $\theta$ by maximizing the marginal likelihood $p(\by_\tau | \bX)$ of observed labels in \eqref{eq-overall-obj}. In prediction, we predict unknown labels with the conditional $p(\by_\kappa | \by_\tau, \bX)$. We stress that we want to account for correlations within $\by_{\kappa}$.

We first suggest that for this per-node classification task, our NMM model should be as good or better than standard GNN classifiers.
Consider the special case of our NMM, if we enforce that each $\bL_i$ is set as $\bL_{ii} = 1$ and $\bL_{ij} = 0$ for $j \neq i$ by setting a large value to $\gamma$ in \eqref{eq-GNN-construction-of-alpha-L}, then all node labels $y_i$ are conditionally independent given $\bX$, and the NMM would be equivalent to a standard GNN model.
With this analysis, we are certain that our NMM, when $L$ is flexible (not constrained to enforce independence), will always match or beat a GNN classifier in terms of training error.

Next, we describe our prediction of unknown labels given known labels for node classification with the conditional $p(\by_\kappa|\by_\tau, \bX)$, where $\kappa$ contains nodes with unknown labels. The conditional can be computed as follows (we omit $\bX$ in notation below for readability): 
\begin{multline}
    p(\by_\kappa | \by_\tau)
    = \sum_{\bc_\tau} p(\bc_\tau | \by_\tau) p(\by_\kappa | \bc_\tau, \by_\tau)
    \\ \quad = \sum_{\bc_\tau} p(\bc_\tau | \by_\tau) \int p(\by_\kappa |\bz_{n(\kappa)}) p(\bz_{n(\kappa)}|\bc_\tau, \by_\tau) d\bz_{n(\kappa)}.
     \nonumber
\end{multline}
The posterior $p(\bz_{n(\kappa)} |\bc_\tau, \by_\tau)$ is computed using conjugacy and conditional independence such that for each node $j \in n(\kappa)$ we have $p(\bz_j |\bc_\tau, \by_\tau) = \text{Dir}( \balpha'_j )$, with parameter vector $\balpha'_{j} = \balpha_j  + \bs_j(\by_\tau, \bc_{\tau})$, with $\bs_j$ defined in \eqref{eq-joint-y-c}.
The integral above defines $p(\by_\kappa | \bc_\tau, \by_\tau)$ is then computed using the conjugacy again. If $\tau$ is a small set of only a few nodes, the summation can be computed in closed form as in the discussion of exact marginals. Otherwise, the exact calculation of $p(\bc_\tau | \by_\tau)$ is not feasible, and neither is the sum over all $\bc_{\tau}$. We instead approximate the sum using $T$ Monte Carlo samples from $q(\bc_\tau | \by_\tau)$ as derived in \eqref{eq-qcy-form}:
\begin{align}
\tilde{p}(\by_\kappa | \by_\tau) =
	\textstyle  \frac{1}{T} \sum_{t=1}^{T} p(\by_\kappa | \bc_\tau^{(t)}, \by_\tau),
	~~ \bc_\tau^{(t)} \sim q(\bc_\tau | \by_\tau).
	\nonumber
\end{align}
When $\kappa$ is small, we check all configurations of $\by_\kappa$ and use the mode of the approximate conditional as the prediction.
When $\kappa$ is large, we predict labels in $\by_\kappa$ one by one using each node's marginal mode.
Each subsequent predicted label is then merged into $\by_\tau$ to decide the next label.

\subsection{Time complexity}
In \emph{training}, the extra cost of NMM compared to a GNN is the computation of $q(\bc_\tau | \by_\tau, \balpha, \bL)$ in \eqref{eq-qcy-form} and extra optimization iterations. Getting a sample from \eqref{eq-qcy-form} takes time $O(|E|T)$ with $T$ being the number of samples and $|E|$ the number of edges. The computation of $q(\bc_\tau | \by_\tau, \balpha, \bL)$  should be faster than a GNN when $T$ is small because it does not deal with features and hidden units as the GNN does. In practice, however, an optimal implementation would need substantial effort for parallelization. In our experiment, we have prioritized a simple implementation without parallelization.

In \emph{prediction} for a single node's label, the runtime cost of our NMM is $O(C D)$ given the GNN output, with number of classes $C$ and maximum degree $D$. We assume $D$ is small ($D \ll N$), so prediction time is not burdensome compared to the cost of producing GNN's prediction  output. Runtime that is linear in neighborhood size is the most affordable complexity one can have when allowing correlations from all neighbors.  

\section{Approximating Other Distributions with the Neighbor Mixture Model}
\label{sec:approx_other_distrib}
As a distribution with easy marginal distributions over nodes, NMM can also approximate another distribution $p^{*}(\by)$ over node labels $\by$, such as an MRF. We only assume $p^*(\by)$ to be an \textit{unnormalized} distribution and to have the ability to evaluate its unnormalized log likelihood $\log p^*(\by)$. 

We seek to minimize the KL-divergence from our model's $p(\by|\bX; \theta)$ to the target $p^*(\by)$ by optimizing our parameters $\theta$. 
The input features $\bX$ can contain information about the target graph and the inference problem, e.g.~we can put MRF potential parameters into $\bX$. If $p^*(\by)$ uses features in the graph (e.g. CRF), we can also put these features to $\bX$. We propose the approximation as
\begin{multline}
\min_{\theta} \KL{p(\by | \bX; \theta)}{p^{*}(\by)}
\\ = \min_{\theta} \E{p}{\log p(\by | \bX; \theta) - \log p^{*}(\by)}.
\end{multline}
The expectation of the first term,  $\E{p}{\log p(\by |\bX; \theta)}$, is intractable as discussed in previous sections. Here we have an upper bound for this term in the spirit of hierarchical variational inference (HVI) \citep{ranganath2016, louizos2017}. The actual optimization problem becomes:  
\begin{multline}
   \min_{\theta}  \E{p}{\log p(\by, \bc | \bX; \theta) - \log q(\bc|\by, \bX; \theta) - \log p^*(\by)} \\ \ge \KL{p(\by | \bX; \theta)}{p^{*}(\by)}. \label{eq-kl-upper}
\end{multline}
Here we use $q(\bc | \by, \bX; \theta) =  q(\bc | \by, (\balpha, \bL) \gets f(\bX; \theta))$ as defined in \eqref{eq-qcy-form} by setting $\tau = V$. Then we can use Monte Carlo samples to estimate this bound. The gradients of the objective with respect to parameters $\theta$ are again estimated by the REINFORCE estimator.

We want to emphasize that using the NMM as the approximating family will have better KL divergence than mean-field distributions in general. If we fix every $\bL_i$ to a one-hot vector such that $p(c_i = i) = 1$, then both $p(\bc | \by, \bX)$ and $q(\bc | \by)$ also become deterministic distributions.
In this special case, the NMM is a mean-field distribution and the upper bound in \eqref{eq-kl-upper} is exactly the KL-divergence of the mean-field distribution. By optimizing $\bL_i$-s, we can easily improve the upper bound to be smaller than the mean-field KL-divergence, and thus the underlying KL-divergence will be even smaller.

For tasks where the target contains both labels and latent variables (e.g. \citet{mehta2019}), which we write as $p^*(\bz, \by)$, we can use the NMM distribution $p(\bz, \by | \bX)$ to approximate the target distribution: 
\begin{align}
\min_{\theta} \E{p(\bz, \by | \bX)}{\log p(\bz, \by | \bX) - \log p^{*}(\bz, \by)}.
    \label{eq-inf-qzy}
\end{align}
Here we can compute $\log p(\bz, \by | \bX)$ efficiently and need no further approximation. We can also use different priors for $\bz$ without requiring conjugacy. Using our NMM distribution is better than assuming mean-field independence because our NMM includes mean-field as a special case, while allowing more flexible correlations if needed.

Our method contributes a way to perform \emph{amortization} across labels that are \emph{dependent} under a target model. It greatly reduces the number of optimization parameters and speeds up inference.
Even in cases when the target model is being dynamically updated (e.g. during model learning), our NMM can be updated alongside in an integrated way.

\section{Experiment}
\begin{table*}[t]
\centering
\scalebox{0.8}{\begin{tabular}{ cccccccccc } 
\hline
\textbf{Dataset} & \textbf{Task} & \textbf{Nodes} & \textbf{Edges} & \textbf{Features} & \textbf{Classes} & \textbf{Training n/e} & \textbf{Validation n/e} & \textbf{Test n/e} & \textbf{Label rate n/e}\\
\hline
Cora & NC/LP & 2708 & 5429 & 1433 & 7 & 140/4616 & 500/271 & 1000/ & 0.05/0.85 \\
Citeseer & NC/LP & 3327 & 4732 & 3703 & 6 & 120/4023 & 500/236 & 1000/473 & 0.04/0.85  \\
Pubmed & NC/LP & 19717 & 44338 & 500 & 3 & 60/37689 & 500/2216 & 1000/4433 & $<$0.01/0.85 \\ 
NIPS12 & LP & 2037 & 3134 & - & - & -/2665 & -/156 & -/313 & -/0.85 \\
Yeast & LP & 2361 & 6646 & - & - & -/5650 & -/332 & -/664 & -/0.85 \\
\hline
\end{tabular}}
\caption{Dataset statistics. NC is node classification; LP is link prediction; n/e is node/edge counts.}
\label{exp-datasets}
\end{table*}

We evaluate our NMM on two types of tasks: modeling observed node labels (node classification and denoising) and inference for other models (an MRF and a model for graph generation). Due to limited  space, full details of protocols and hyperparameters are in the appendix.

\subsection{Graph Node Classification}
\begin{table}[t]
    \centering
    \scalebox{0.85}{
    \begin{tabular}{clll}
        \hline
        \textbf{Algorithm} & \textbf{Cora} & \textbf{Citeseer} & \textbf{Pubmed}\\
        \hline
        GCN & 81.3$\pm$0.9 & 71.0$\pm$0.7 & 79.0$\pm$0.4\\
        GAT & 83.0$\pm$0.7 & 72.5$\pm$0.7 & 79.0$\pm$0.3\\
        APPNP & 84.1$\pm$0.9 & 71.7$\pm$0.8 & 79.4$\pm$0.3\\
        \hline
        GMNN & 82.1$\pm$1.4 & 71.4$\pm$1.1 &
        \textbf{80.6$\pm$0.7}\\
        \hline
        NMM-GCN & 84.3$\pm$0.5 \checkmark & 72.0$\pm$0.3 \checkmark & 79.2$\pm$0.2\\
        NMM-GAT & 84.4$\pm$0.2 \checkmark & 73.0$\pm$0.5 & 79.4$\pm$0.2 \checkmark\\
        NMM-APPNP & \textbf{85.9$\pm$0.5} \checkmark & 72.6$\pm$0.3 \checkmark & \textbf{80.4$\pm$0.2} \checkmark\\
        \hline
    \end{tabular}
    }
    \caption{Node classification results on three datasets. Bold numbers indicates the best performance, and the $\checkmark$ mark indicates the combination with NMM outperforms the corresponding baseline.}
    \label{exp-nc}
\end{table}
\begin{table}[]
    \centering
    \scalebox{0.72}{
    \begin{tabular}{cccccccc}
        \hline
        \multirow{2}{*}{\textbf{Algorithm}} & & \textbf{Cora} & & & & \textbf{Citeseer} &\\
        \cline{2-4} \cline{6-8}
        & \textbf{0.2} & \textbf{0.4} & \textbf{0.6} & & \textbf{0.2} & \textbf{0.4} & \textbf{0.6}\\
        \cline{2-4} \cline{6-8}
        GMNN & -1.34 & -1.30 & -1.14 & & -1.62 & -1.53 & -1.41\\
        NMM-GCN & -1.04 & -0.97 & -0.81 & & -1.37 & -1.34 & -1.26\\
        NMM-GAT & -1.01 & \textbf{-0.84} & \textbf{-0.71} & & \textbf{-1.35} & -1.34 & \textbf{-1.25}\\
        NMM-APPNP & \textbf{-0.93} & \textbf{-0.83} & \textbf{-0.70} & & \textbf{-1.33} & \textbf{-1.30} & \textbf{-1.24}\\
        \hline
    \end{tabular}
    }
    \caption{Marginal log-likelihood of label pairs $y_i, y_j$ in the test set across fractions of observed data included in training set (0.2, 0.4, and 0.6). NMM assigns higher heldout probability than the GMNN. Statistically indistinguishable best values are bold. Full table with standard deviations over multiple runs in appendix.}
    \label{exp-pairwisell}
\end{table}

We follow previous literature \citep{kipf2017, velickovic2018, qu2019} and evaluate our methods on three node classification benchmarks: Cora, Citeseer, and Pubmed.  Summary statistics are listed in Table \ref{exp-datasets}. Our data split follows \citet{yanga16}: we select 20 nodes from each class from training. We assess each model's \emph{accuracy} on the test set.

We use three popular graph neural networks as baselines: Graph
Convolutional Network (GCN) \citep{kipf2017}, Graph
Attention Network (GAT) \citep{velickovic2018}, and Approximate Personalized Propagation of Neural Predictions (APPNP) \citep{klicpera2019}. These methods all assume node labels are independent given node features.
We then build our proposed NMM model using each one of these GNNs as the backbone neural network in \eqref{eq-GNN-construction-of-alpha-L}.
We refer to each variant of our method as NMM-GCN, NMM-GAT, and NMM-APPNP.
We also compare against the recent Graph Markov Neural Network (GMNN) \citep{qu2019}, which combines MRFs with GNNs to model node label dependencies. The temperature parameter of GMNN is set to 1.0, leading to a formal probabilistic model as argued in the original paper.

Our implementation is based on the PyTorch Deep Graph Library (DGL) \citep{wang2019}. We reuse the GCN results from \citet{wang2019} to consider the standard deviation. To fairly compare with GAT, we take the best results from either the original paper \citep{velickovic2018} or our rerun using DGL and a larger GAT model.

Table \ref{exp-nc} gives the mean node classification accuracy   as well as standard deviation over five random repeats. Results of different algorithms are compared via a $t$-test with $p<0.05$. The best results and any statistically indistinguishable from the best are bold. We see that our model combined with APPNP achieves the best performance in general. Other combinations also outperform the GMNN on two out of three datasets. For all three backbone GNNs, integrating it with NMM leads to significantly improved performances (marked as $\checkmark$) on at least two of the three datasets. This result confirms that incorporating neighborhood label dependencies into GNN models is beneficial.

\begin{figure}[t]
\begin{minipage}[b]{0.2\textwidth}
    \centering
      \includegraphics[width=1.2\linewidth]{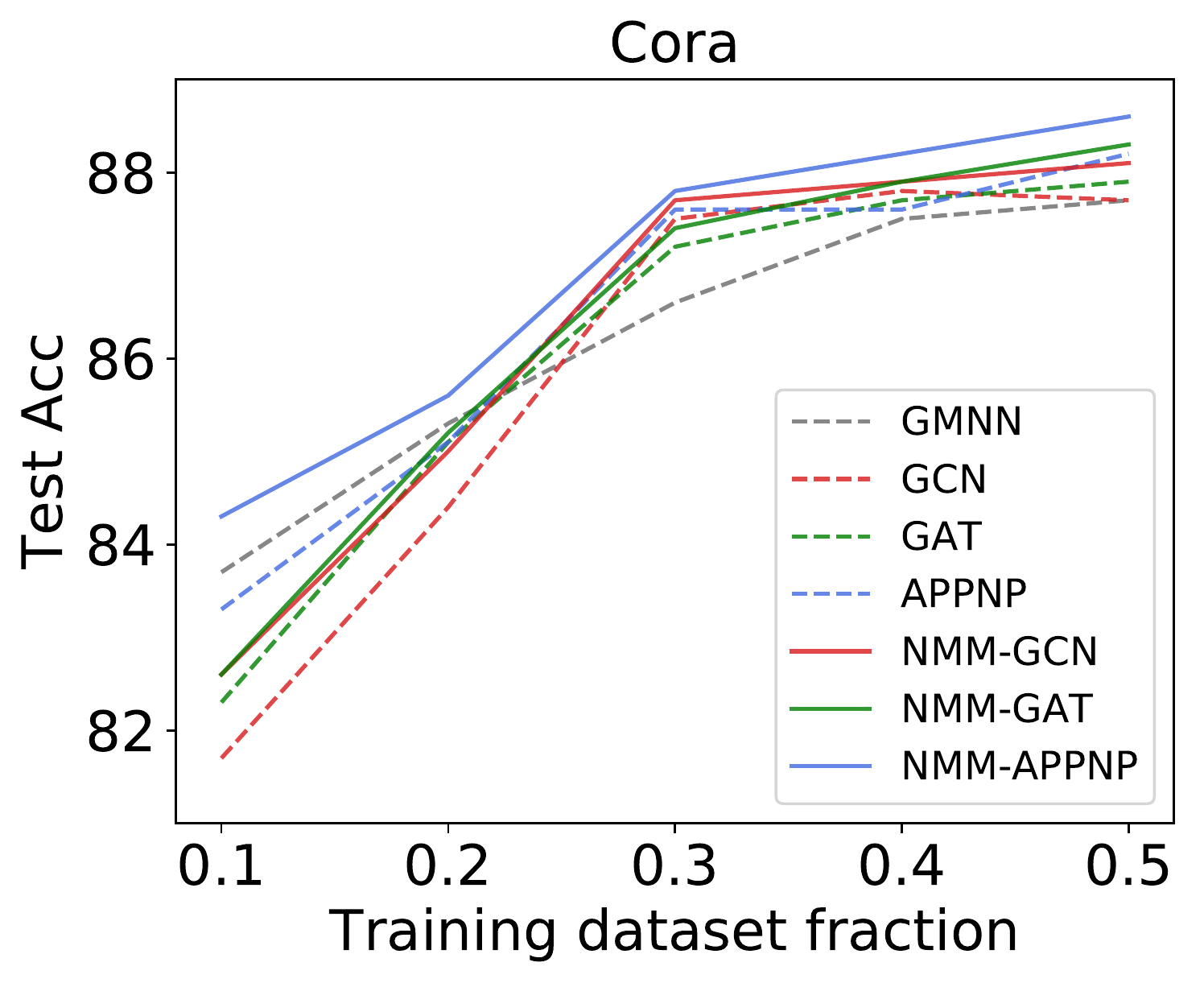}
\end{minipage}
\hspace{5mm}
\begin{minipage}[b]{0.2\textwidth}
    \centering
      \includegraphics[width=1.2\linewidth]{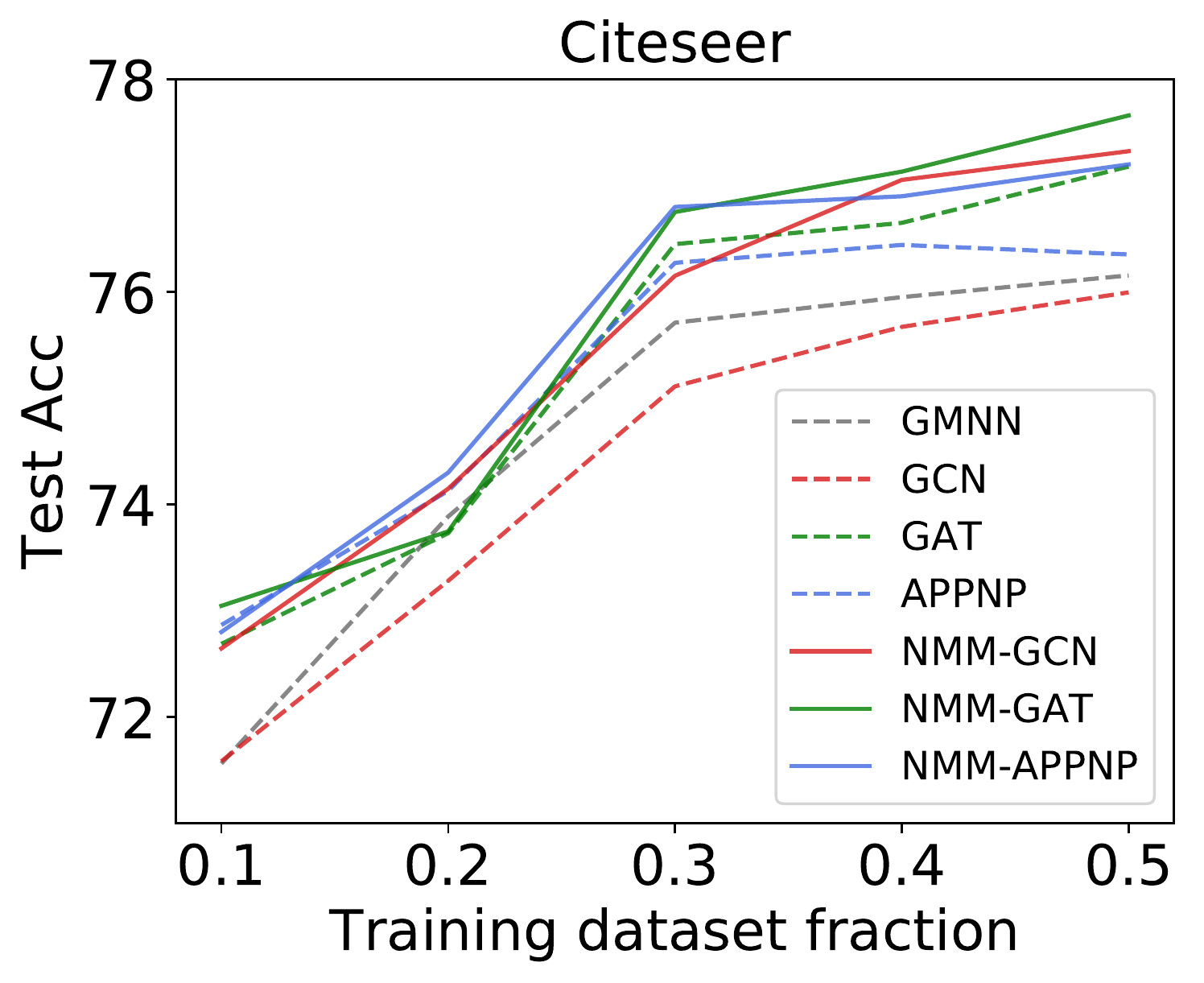}
\end{minipage}
\caption{Classification accuracy on heldout test nodes as training set size increases. Training ratio at 0.5 results in the highest test accuracy. This result is higher than the results in Table \ref{exp-nc} where we used a smaller training ratio.}
\label{exp-nc-increase-train}
\end{figure}

To assess robustness, we test all algorithms at different training set sizes (varying the \emph{fraction} of nodes included in the training set).
We vary this fraction from 0.1 to 0.5 while fixing validation and test fractions to 0.2 and 0.3. Figure \ref{exp-nc-increase-train} shows test accuracy of all algorithms on Cora and Citeseer datasets. Every NMM-GNN combination (solid lines) outperforms its corresponding baseline GNN (same color, dashed). All NMM-GNN models also deliver better performance than GMNN. 

Accurate estimation of probabilistic dependencies is an important aspect of probabilistic models. Node classification accuracy only reflects marginal probabilities of single nodes. To better compare estimated correlations between nodes, we examine the \emph{pairwise marginal log likelihood} (PLL) reported from both GMNN and NMM models. Formally, we compute PLL as the average log-likelihood per edge: $ PLL = \frac{1}{|E_{test}|} \sum_{(i, j) \in E_{test}} \log p(y_{i}, y_{j})$.
We randomly choose an edge set $E_{test}$ such that each edge has two incident nodes from the test node set. We vary the training ratio from 0.2 to 0.6 and fix the validation ratio (0.1) and test ratio (0.3). Table \ref{exp-pairwisell} shows that our NMM achieves better predictions of pairwise probabilities than GMNN, suggesting that our NMM can better capture correlations between connected nodes.

\begin{figure}[t]
\begin{minipage}[b]{0.2\textwidth}
    \centering
      \includegraphics[width=1.2\linewidth]{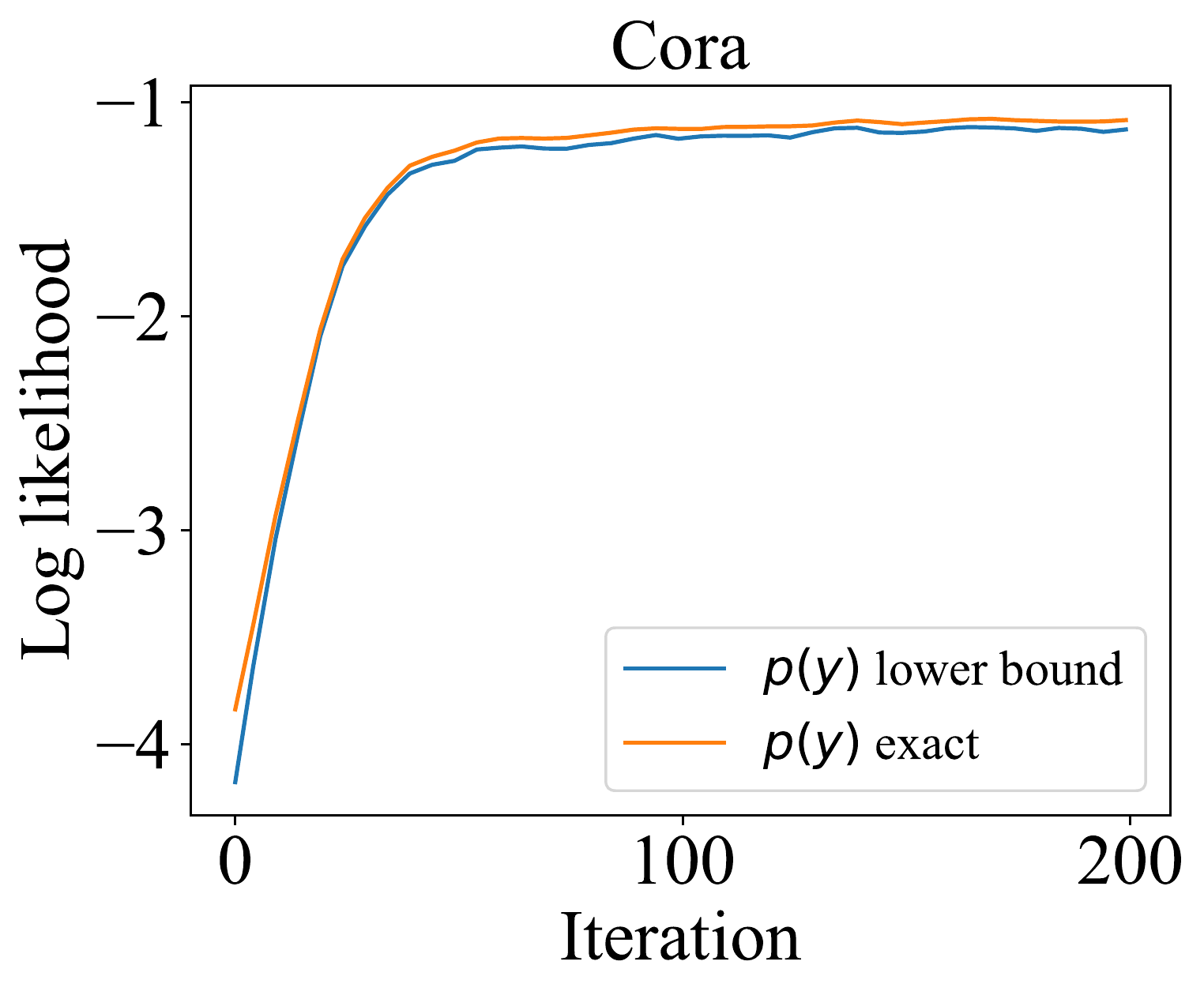}
\end{minipage}
\hspace{5mm}
\begin{minipage}[b]{0.2\textwidth}
    \centering
      \includegraphics[width=1.2\linewidth]{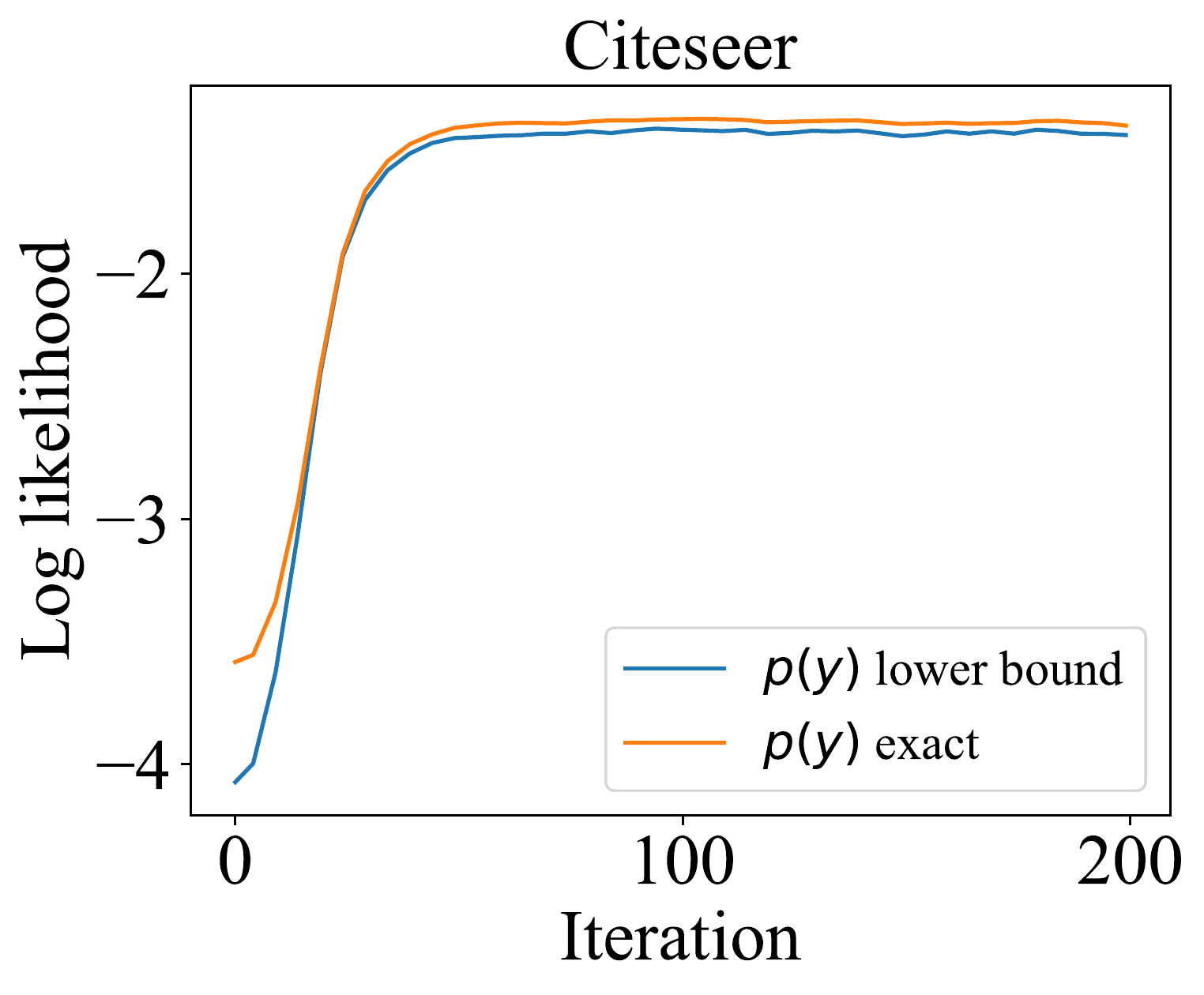}
\end{minipage}
\caption{Marginal log-likelihood of label pairs $y_i, y_j$ in the test set over many training iterations, using NMM-GCN with 0.2 data fraction. Our scalable bound is a tight approximation of the expensive exact.}
\label{exp-lb}
\end{figure}

Finally, we also inspect the quality of the approximation by the variational distribution in \eqref{eq-learn-obj}. We do so on the pairwise marginals $p(y_{i}, y_{j}), (i, j) \in E_{test}$. Figure \ref{exp-lb} compares the variational lower bound against the exact log marginal. The bound is quite tight, which is a strong evidence that the variational distribution $q(\bc | \by, \balpha, \bL)$ is an accurate approximation of the true posterior $p(\bc | \by, \balpha, \bL)$.

\subsection{Image Denoising}

\begin{table*}[!t]
\centering
\scalebox{0.95}{\begin{tabular}{ cccccc } 
\hline
\textbf{Algorithm} & \textbf{NIPS12} & \textbf{Yeast} & \textbf{Cora} & \textbf{Citeseer} & \textbf{Pubmed} \\
\hline
DGLFRM & 88.66 $\pm$ 0.56 & 84.03 $\pm$ 0.69 & 93.44 $\pm$ 0.40 & 94.31 $\pm$ 0.36 & \textbf{96.47 $\pm$ 0.18} \\
\hline
NMM-DGLFRM & \textbf{89.82 $\pm$ 0.64} & \textbf{84.92 $\pm$ 0.51} & \textbf{94.16 $\pm$ 0.47} & \textbf{94.75 $\pm$ 0.38} & \textbf{96.45 $\pm$ 0.14} \\
\hline
\end{tabular}}
\caption{Link prediction results, measured by Average Precision (AP, higher is better).}
\label{exp-sbm-ap}
\end{table*}

\begin{figure}[t]
	\setlength{\tabcolsep}{1.0mm}
	\begin{tabular}{c c c c}
        \includegraphics[width=1.6cm]{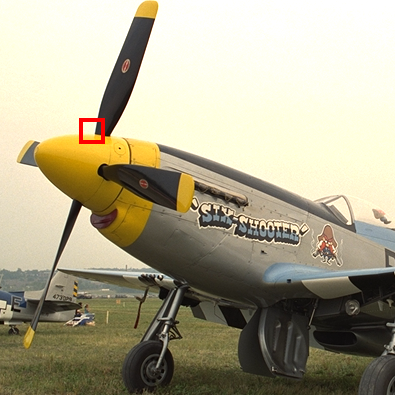}
		& \includegraphics[width=1.6cm]{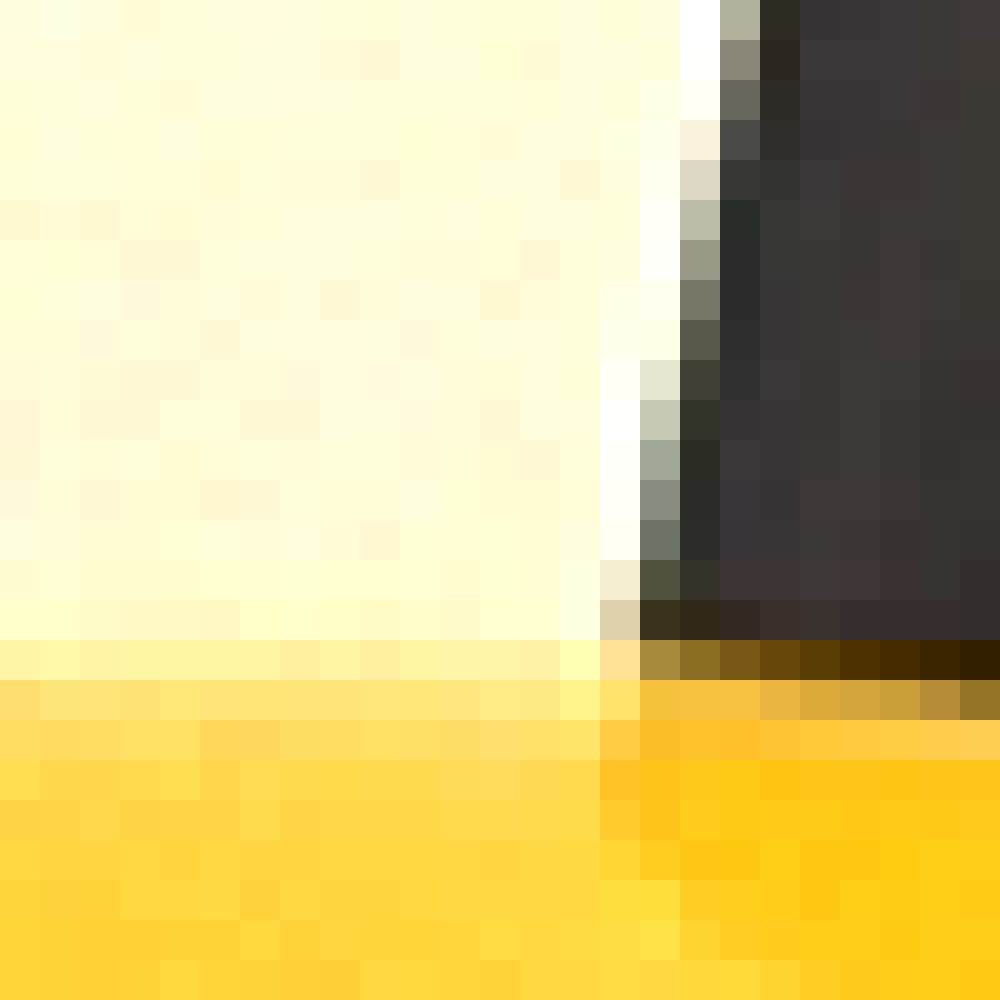}
        & \includegraphics[width=1.6cm]{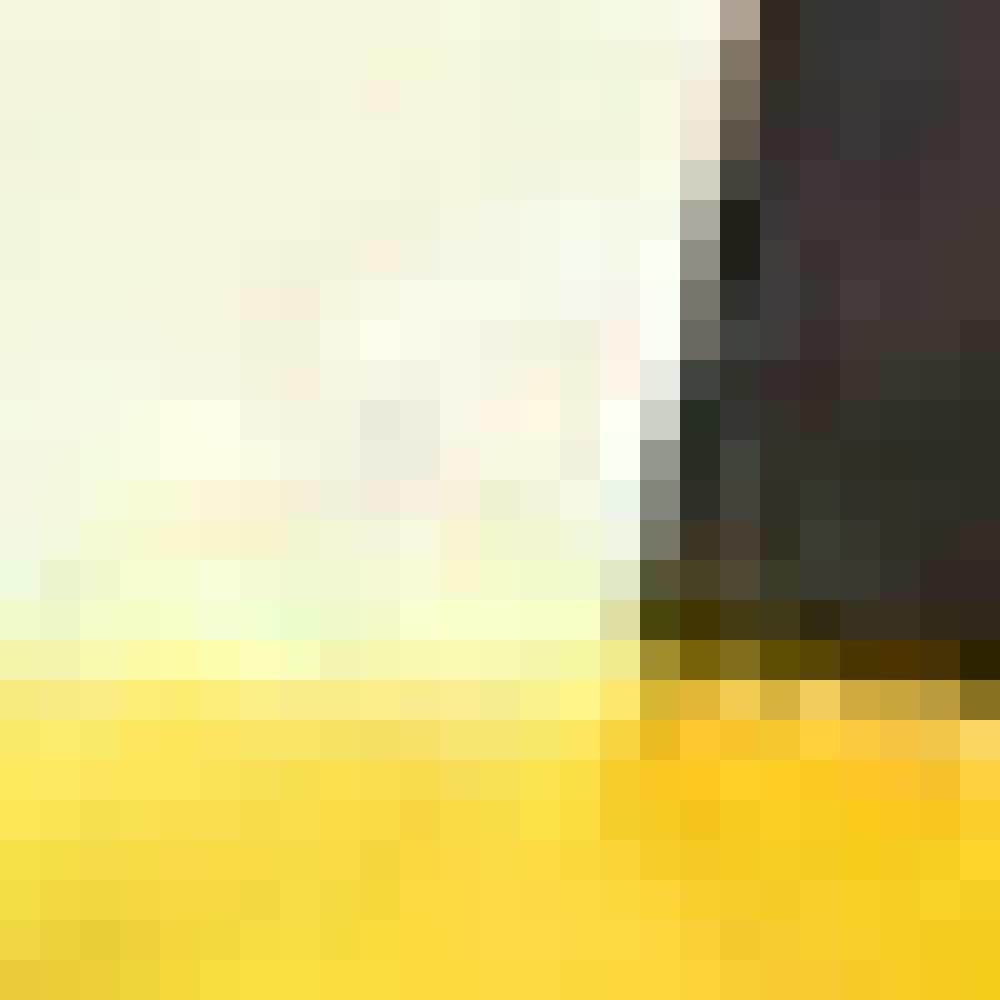}
        & \includegraphics[width=1.6cm]{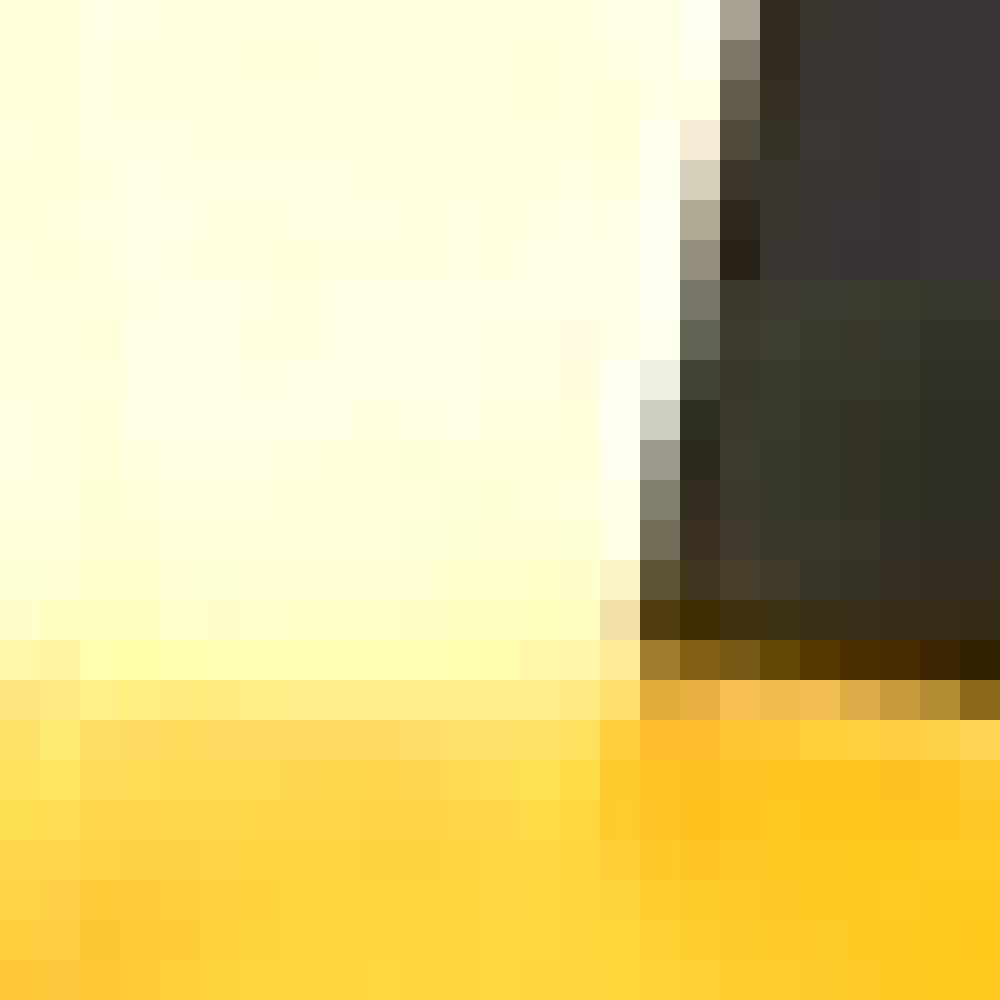}\\
		Original Image & Clean Patch & Baseline & NMM 
	\end{tabular}
	\caption{Image denoising result. NMM recovers the sky and sharp propeller edge better than the baseline.}
	\label{exp-denoising-showcase}
\end{figure}
\begin{table}[t]
    \centering
    \scalebox{0.95}{
    \begin{tabular}{ccc}
        \hline
        \textbf{Method} & \textbf{BSD300} & \textbf{Kodak}\\
        \hline
        Baseline & 30.28 & 31.07 \\
        NMM & \textbf{30.33} & \textbf{31.55} \\
        \hline
    \end{tabular}}
    \captionof{table}{PSNR (the larger the better) performance for image denoising on BSD300 testing set and Kodak.}
    \label{exp-denoising-n2c}
\end{table}
\begin{table}[t]
    \centering
    \scalebox{0.95}{
\setlength{\tabcolsep}{1mm}  
    \begin{tabular}{ cccc } 
    \hline
     & {\small \textbf{MF}} & \textbf{\small NMM}-$\balpha,\bL$ & \textbf{\small NMM-CNN} \\
    \hline
    KL/UB & 3179 & \textbf{3080} & 3105 \\
    \hline
    \end{tabular}}
    \caption{KL divergence or upper bounds on KL for methods that approximate an MRF (lower is better).}
    \label{exp-mrf-kl}
\end{table}

We also evaluate NMM model on an image denoising task. Recent methods use U-Nets \citep{ronneberger} for denoising \citep{mao2016, lehtinen2018}, which do not model pixel correlations. In this experiment, we use NMM to model pixel correlations and put a U-Net as the backbone of the NMM. We treat pixel values as discrete labels, and each pixel's neighborhood is the $3\times 3$ area around it.  

The baseline U-Net and the NMM are trained with noisy images as inputs and clean images as targets. The training set is the BSD300 \citep{martin2001} training set. Due to resource limitations, both models are trained on $32 \times 32$ patches, not full images. The trained models are evaluated on the BSD300 test set and the Kodak dataset\footnote{\url{http://r0k.us/graphics/kodak/}}. Following \citet{lehtinen2018}, we inject Gaussian noise to obtain training (std. dev. $\sigma \in [0, 50]$) and test ($\sigma =25$) images. 

Table \ref{exp-denoising-n2c} shows the  peak Signal-to-Noise Ratio (PSNR) of denoised images by different methods (the larger the better), and Figure \ref{exp-denoising-showcase} shows a typical example on which the NMM outperforms the baseline. Note that the PSNR is directly computed from the per-pixel Mean Squared Error (MSE) and the U-Net directly minimizes MSE during training, so the evaluation measure favors the U-Net. Nevertheless, our NMM still achieves better PSNR. The main reason is that images of the natural world have strong correlations between neighboring pixels, and NMM can better capture such correlations. In Figure \ref{exp-denoising-showcase}, the NMM achieves qualitatively better denoising results in smooth areas while the attention mechanism avoids blurring to keep edges sharp.

\subsection{Approximating other distributions}

\textbf{Approximating an MRF.}
We first use NMM to approximate inference for an MRF, as described in Sec.~\ref{sec:approx_other_distrib}. We create an Ising model on a $128 \times 128$ grid \citep[section 21.3.2]{murphy2012}. As a baseline, we approximate the MRF with the mean-field distribution (MF). We use two NMMs to approximate the MRF: one that directly optimizes $\balpha$ and $\bL$ (NMM-$\balpha,\bL$), and one that optimizes a CNN to compute $\balpha$ and $\bL$ (NMM-CNN).  The input to the CNN is a grid graph with all MRF edge potential and node potential parameters collated at grid nodes. 
For the two NMMs, we can only compute an upper bound in \eqref{eq-kl-upper}. Table \ref{exp-mrf-kl} shows that the upper bounds of the two NMMs achieve noticeably better approximations (lower KL) than the mean-field baseline. The relatively small difference (<1\%) between NMM-$\balpha,\bL$ and NMM-CNN also indicates that the CNN as an inference network is strong enough to get good inference results despite its amortized parameterization.

\textbf{Approximating deep generative graph models with latent variables.}
Next, we use our NMM to do inference for the Deep Generative Latent Feature Relational Model (DGLFRM) \citep{mehta2019}. Formally, DGLFRM itself defines a deep generative stochastic blockmodel with random variables $\by_i$ and $\bz_i$\footnote{The original paper uses notation $p(\bv, \bb)$, which maps to our notation as $\bz=\bv$, $\by=\bb$. We omit notation of their variable $\br$ for conciseness, as we keep the same parameterization of $\br$ as the original work.}.
Binary vector $\by_i \in \{0, 1\}^K$ indicates node $i$'s memberships in $K$ communities, and latent variable $\bz_i$ serves as stick-breaking construction of the Indian Buffet Process (IBP) for $\by_i$ \citep{teh2007}.
DGLFRM pursues variational inference with a structured mean-field approximate posterior:
\begin{align}
    q(\bz, \by) = \textstyle \prod_{k=1}^K \prod_{i=1}^N q(z_{ik})q\left(y_{ik}|\pi_{ik}(\bz_i)\right). \nonumber
\end{align}
Here, $z_{ik} \sim \mathrm{Beta}(a_k, b_k)$, $y_{ik} \sim \mathrm{Bernoulli}\left(\pi_{ik}(\bz_i)\right)$, and $\pi_{ik}(\bz_i) =\prod_{j=1}^{k} z_{ij}$.
The probability vector $\bm{\pi}_i(\bz_i)$  is a transformation of \textit{i.i.d.} random variables $\bz_i$.

We adapt our NMM to approximate a target posterior distribution like $q(\bz, \by)$ with slight modifications: instead of sampling $\by_i$ with probability $\mathbold{\pi}_i$, our NMM still introduces a random variable $c_i$ for node $i$, and then we sample $\by_i$ from $\mathbold{\pi}_{c_i}$. As in the original NMM definition, the indicator variable $c_i$ introduces membership correlations between neighboring nodes, which the original inference above cannot do.

We trained two models: DGLFRM with its original inference, and our NMM-DGLFRM. We evaluate link prediction over five datasets: NIPS12, Yeast, Cora, Citeseer, and Pubmed, following \citet{mehta2019}. For each dataset, 10\% and 5\% of edges are held out as test and validation sets. We report Average Precision (AP) as the evaluation metric. 

Table \ref{exp-sbm-ap} shows that NMM-DGLFRM outperforms the baseline model on four out of five datasets. We also observe that our model achieves higher variational lower bound than the baseline DGLFRM (see the appendix). Since neighboring nodes often share similar memberships, information sharing between neighbors clearly improves the approximation of the true posterior. Thus, our NMM method's better inference has improved the model's overall  probabilistic representation of graph data (e.g. better link prediction).

\section{Conclusion}
In this paper, we have presented a new model, the Neighbor Mixture Model (NMM), which captures probabilistic correlations among labels arranged in a graph. Parameterized by a GNN, the model learns informative representations from input features. It enables scalable computation via a tractable variational lower bound that requires no additional free parameters. The model can also serve as an approximate distribution to enable scalable inference for other models. The NMM's high-quality performance across multiple tasks indicates promising ability to model node correlations at scale, while easily integrating with many other models.

\bibliographystyle{apalike}
\bibliography{spatial}

\appendix
\onecolumn
\section{Calculating of the Joint Probability of $\by$ and $\bc$}

The closed-form of $p(\by_\tau, \bc_\tau| \balpha, \bL)$ is needed for marginalization (Sec. 3.2 of main paper), variational lower bound (Sec. 3.3), and conditional probability (Sec. 3.6). Here we give a detailed derivation. The general idea is to leverage the Dirichlet-categorical conjugacy to do the computation. 
\begin{align}
    p(\by_\tau, \bc_{\tau} | \balpha, \bL) &= \int_{\bz_{n(\tau)}} p(\by_\tau, \bc_{\tau}, \bz_{n(\tau)} | \balpha, \bL) \nonumber \\
    &= \int_{\bz_{n(\tau)}} p(\bz_{n(\tau)} | \balpha) p(\bc_{\tau} | \bL) p(\by_\tau | \bc_\tau, \bz_{n(\tau)}) \nonumber \\
    &= p(\bc_{\tau} | \bL) \int_{\bz_{n(\tau)}} p(\bz_{n(\tau)} | \balpha) \prod_{i\in \tau} p(y_i | c_i, \bz_{n(i)}) \nonumber \\
    &= \prod_{i \in \tau} L_{i, c_i} \int_{\bz_{n(\tau)}} \prod_{j \in n(\tau)} \text{Dir}(\bz_j | \balpha_j) \prod_{i\in \tau} z_{c_i, y_i} \nonumber \\
    &= \prod_{i \in \tau} L_{i, c_i}  \prod_{j \in n(\tau)} \frac{\mathrm{B}(\balpha_j + \bs_j(\by_\tau, \bc_\tau))}{\mathrm{B}(\balpha_j)} \nonumber \\
    &= \prod_{i \in \tau} L_{i, c_i}  \prod_{j \in \bc_\tau} \frac{\mathrm{B}(\balpha_j + \bs_j(\by_\tau, \bc_\tau))}{\mathrm{B}(\balpha_j)}.
\end{align}
Here the integral is computed from the  Dirichelet-multinomial conjugacy. The vector $\bs_j(\by_\tau, \bc_\tau) = \sum_{i\in \tau: c_i = j} \mathrm{onehot}(y_i)$, and $\mathrm{B}(\balpha) = \frac{\Gamma(\alpha_1) \Gamma(\alpha_2) \ldots \Gamma(\alpha_C)}{\Gamma(\sum_k \alpha_k)}$ denotes the multivariate Beta function.

\section{Graph Node Classification: Details and Additional Results}

\subsection{Experimental Details}

In node classification, we have used several possible GNN architectures (GCN, GAT, and APPNP) to construct $\balpha$ and $\bL$. In the paper, we described a single function  $f(\bX; \theta)$ that used a GNN to compute hidden representation vectors $(\bu_i, \bv_i)_{i \in V}$ for every node $i$, and then compute $\balpha$ from $(\bu_i)_{i \in V}$ and compute $\bL$ from $(\bv_i)_{i \in V}$. In the actual implementation, we have used two separate graph neural networks to compute $(\bu_i)_{i \in V}$ and $(\bv_i)_{i \in V}$. Essentially, this means our function $f$ in practice does not share structure between computation of $\bu_i$ and $\bv_i$, but nothing prevents trying other implementations in the future.

We now review our concrete architectures for computing $(\bu_i)_{i \in V}$ with each possible GNN architecture. For GCN, we use a two layer GCN with 16 hidden nodes as in \citet{kipf2017}. For GAT, we use a two layer GAT with 8 attention heads in both layers and 16 hidden nodes (slightly larger than the 8 hidden nodes used in \citet{velickovic2018}). Finally, for APPNP we use a two layer APPNP with 0.1 teleport probability, 10 propagation steps, and 64 hidden nodes as in \citet{klicpera2019}.

Similarly, we can compute $(\bv_i)_{i \in V}$ with each possible GNN architecture. Our GCN uses a one layer GCN (64 hidden nodes). Our GAT uses a one layer GAT (32 hidden nodes and 8 heads). APPNP uses the same architectures to compute $\bu_i$-s and $\bv_i$-s, except that the output dimension for a $\bv_i$ is 32. Experiments are conducted in a Tesla V100 GPU cluster.

\textbf{Hyperparameters.}
Hyperparameters are selected by checking performance on a fixed validation set. Models are trained on a maximum of 200 epochs, stopping early if validation performance has not improved over a contiguous window of 100 epochs. We use Adam with a step size pool $\{0.01, 0.03, 0.05, 0.07, 0.1\}$. We apply $L_2$ regularization with $\gamma=0.005$ and dropout rate from $\{0.5, 0.7\}$. The output activation for $\balpha$ inference network is validated from $\{\softplus(\cdot), \mathrm{square(\cdot)}\} + 1.0$, to ensure the $\balpha$ defines valid Dirichlet densities with a unique mode. In our experiment, we observe that $\softplus(\cdot)$ is better at stabilizing the training process while $\mathrm{square(\cdot)}$ enables faster training, especially when the output logits of the $\balpha$ inference network are negative. For all node classification experiments, to compute $\bL$ we set the ``self-attention'' scalar $\gamma$ to be 0.0, and cosine similarity scale $\sigma^2$ is optimized with an initial value 1.0. 

\textbf{Baselines.} The GMNN model has a ``temperature'' parameter to sharpen the variational distribution. This ad-hoc parameter leads to slightly better performance but very poor probability estimation. Throughout our experiments, we have tested the model with the temperature parameter setting to 1.0, which yields ``standard'' variational inference.

\subsection{Additional Results: Runtime and Convergence Analysis}
\begin{table}[t]
    \centering
    \begin{tabular}{cccc}
    \hline
    \textbf{Algorithm} & \textbf{Cora}  & \textbf{Citeseer} & \textbf{Pubmed} \\
    \hline
    GCN       & 0.01s & 0.02s      & 0.04s\\
    GMNN      & 0.03s & 0.07s      & 0.14s\\
    NMM-GCN   & 0.04s & 0.03s      & 0.05s\\
    \hline
    \end{tabular}
    \caption{Training time per epoch. Each epoch has a forward pass, a loss computation, and a backward pass.}
    \label{sup-exp-runtime}
\end{table}
\begin{figure}[t]
    \begin{minipage}[b]{0.25\textwidth}
        \centering
         \includegraphics[width=1.2\linewidth]{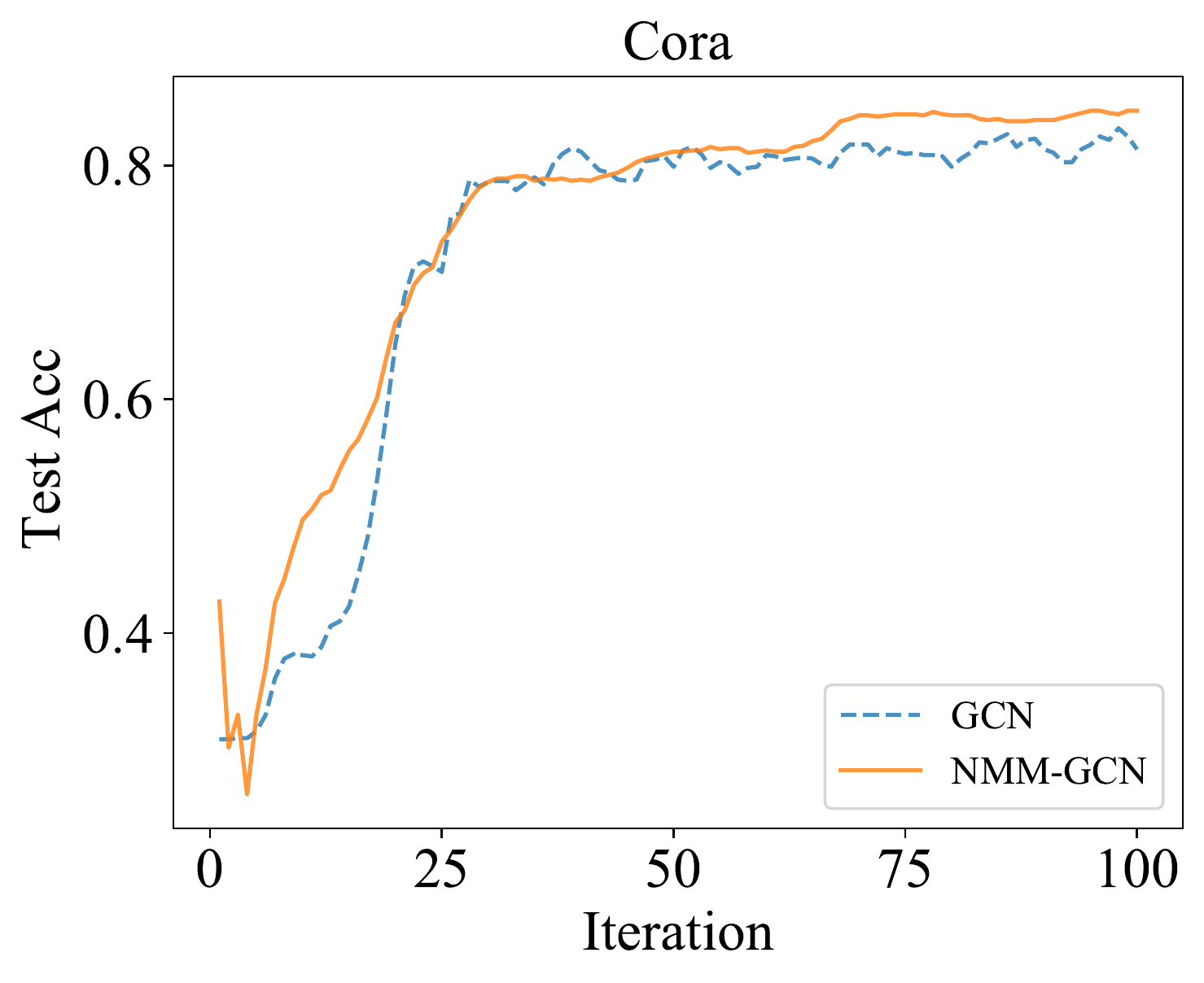}
    \end{minipage}
    \hspace{10mm}
    \begin{minipage}[b]{0.25\textwidth}
        \centering
        \includegraphics[width=1.2\linewidth]{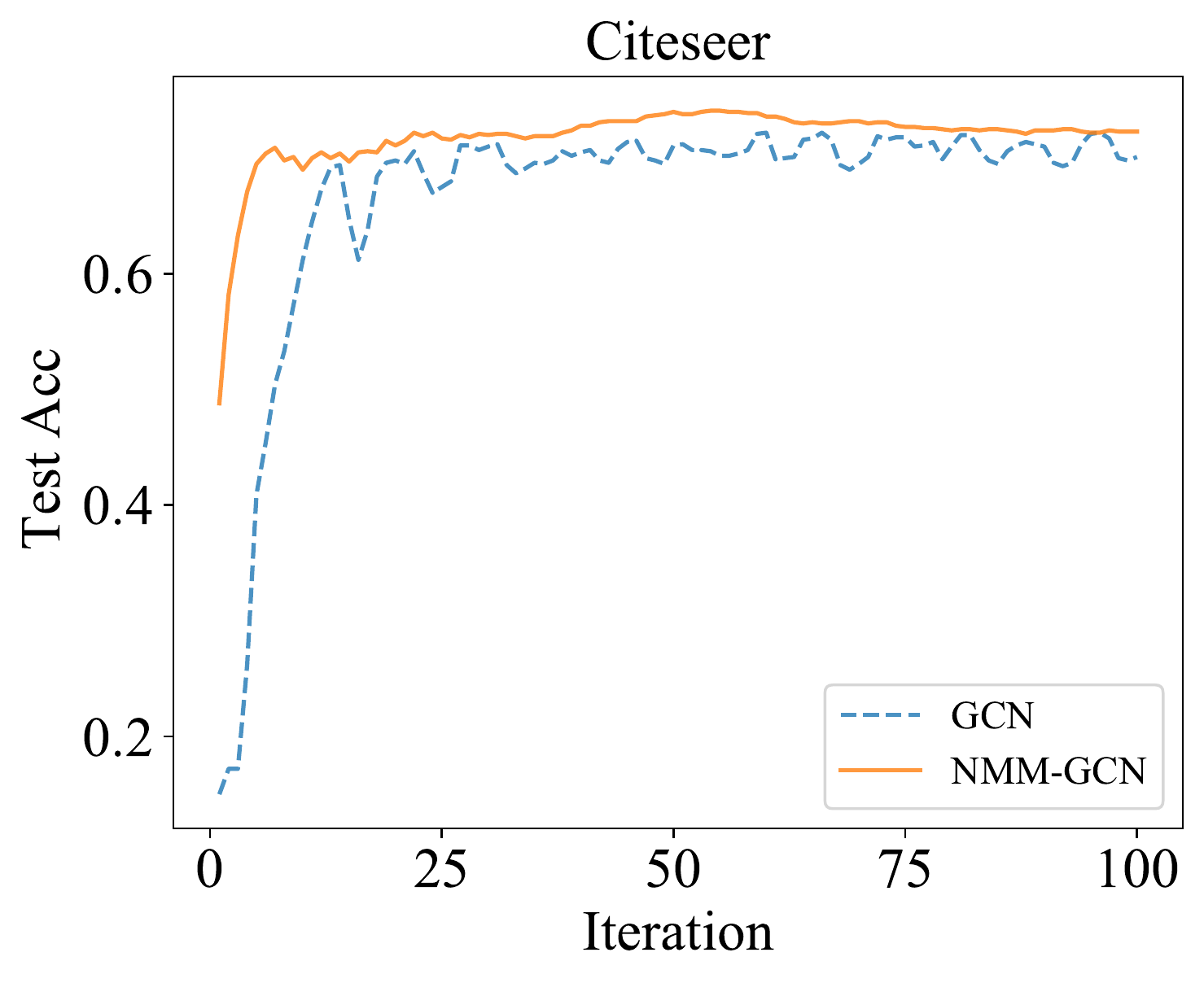}
    \end{minipage}
    \hspace{10mm}
    \begin{minipage}[b]{0.25\textwidth}
        \centering
        \includegraphics[width=1.2\linewidth]{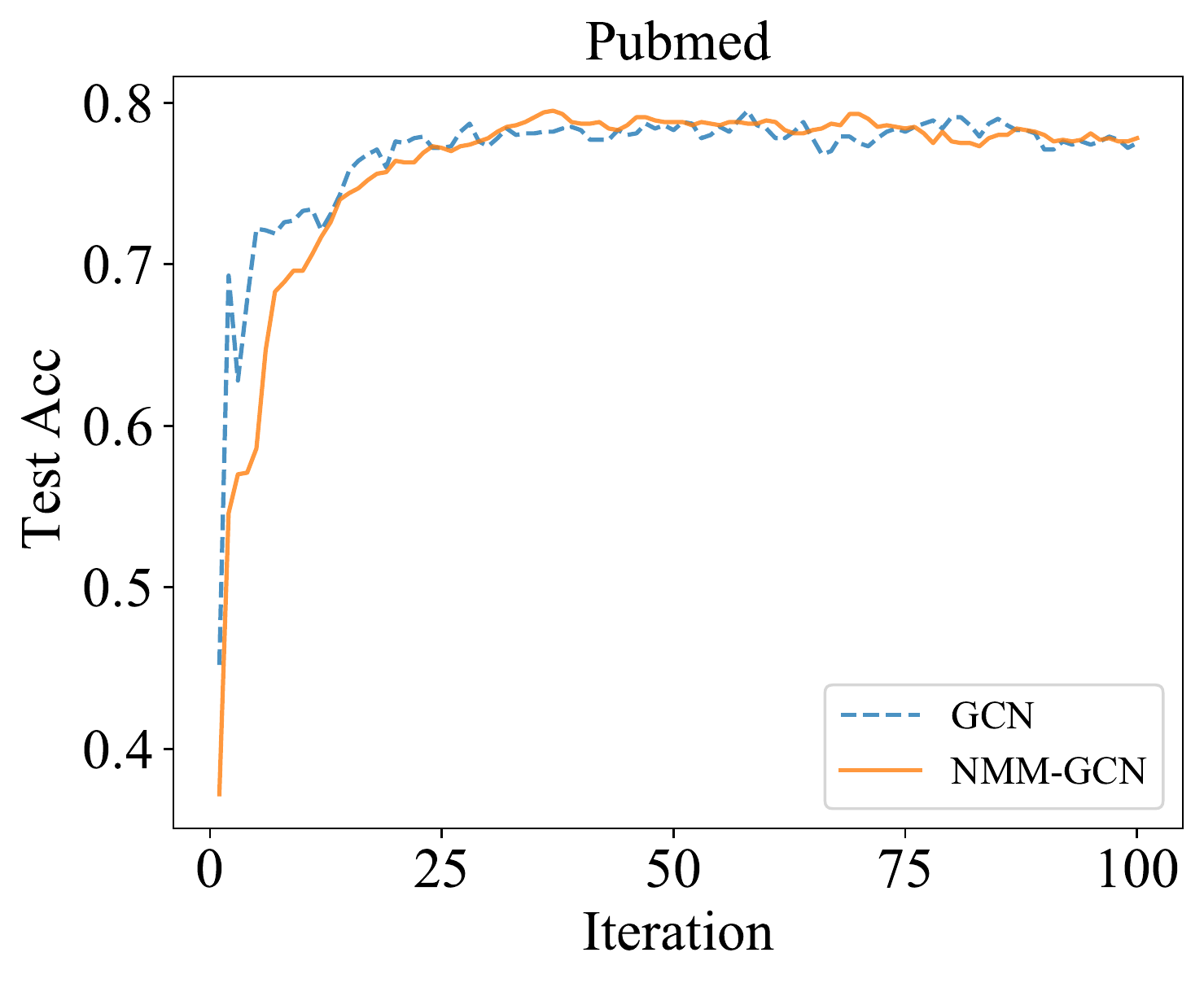}
    \end{minipage}
    \caption{Convergence curves of GCN and NMM-GCN.}
    \label{sup-exp-convergence}
\end{figure}
\parhead{Runtime.} In addition to time complexity (Sec. 3.7 of main paper), we report mean training
time per epoch for 100 epochs to probe practical runtime \citep{kipf2017}. Table \ref{sup-exp-runtime} summarizes the results. NMM-GCN introduces minor runtime overhead its backbone GCN. NMM-GCN is faster than GMNN on Citeseer and Pubmed. Similar results are observed when using other backbones for NMM (GAT and APPNP)

\parhead{Convergence Analysis.} Figure \ref{sup-exp-convergence} shows the convergence curves of GCN and NMM-GCN. We see NMM-GCN has similar convergence behavior as its backbone GCN. We did not include convergence curve of GMNN, as GMNN has a pre-training stage beforehand.

\subsection{Additional Results: Pairwise Probability Assessment}
Table \ref{sup-exp-pairwisell} gives full results for pairwise log likelihood (PLL, Table 3 of main paper). Mean PLL and standard deviation are reported over five random runs. The results indicate that the NMM has better calibrated probabilities than GMNN.
\begin{table}[t]
    \centering
    \scalebox{0.9}{
    \begin{tabular}{cccccccc}
        \hline
        \multirow{2}{*}{\textbf{Algorithm}} & & \textbf{Cora} & & & & \textbf{Citeseer} &\\
        \cline{2-4} \cline{6-8}
        & \textbf{0.2} & \textbf{0.4} & \textbf{0.6} & & \textbf{0.2} & \textbf{0.4} & \textbf{0.6}\\
        \cline{2-4} \cline{6-8}
        GMNN & -1.34 $\pm$ 0.06 & -1.30  $\pm$ 0.09 & -1.14 $\pm$ 0.04 & & -1.62 $\pm$ 0.08 & -1.53 $\pm$ 0.05 & -1.41 $\pm$ 0.04\\
        NMM-GCN & -1.04 $\pm$ 0.01 & -0.97 $\pm$ 0.08 & -0.81 $\pm$ 0.03 & & -1.37 $\pm$ 0.02 & -1.34 $\pm$ 0.01 & -1.26 $\pm$ 0.01\\
        NMM-GAT & -1.01 $\pm$ 0.01 & \textbf{-0.84 $\pm$ 0.06} & \textbf{-0.71 $\pm$ 0.01} & & \textbf{-1.35 $\pm$ 0.02} & -1.34 $\pm$ 0.02 & \textbf{-1.25 $\pm$ 0.02}\\
        NMM-APPNP & \textbf{-0.93 $\pm$ 0.01} & \textbf{-0.83 $\pm$ 0.04} & \textbf{-0.70 $\pm$ 0.02} & & \textbf{-1.33 $\pm$ 0.02} & \textbf{-1.30 $\pm$ 0.02} & \textbf{-1.24 $\pm$ 0.01}\\
        \hline
    \end{tabular}
    }
    \caption{Marginal log-likelihood of label pairs $y_i, y_j$ in the test set across fractions of observed data included in training set (0.2, 0.4, and 0.6).}
    \label{sup-exp-pairwisell}
\end{table}

\section{Image Denoising: Details}
The baseline U-Net \citep{ronneberger} is trained to fit pixel values. We also use such a network as the backbone of the NMM. We convert a scalar prediction $u_i$ to a positive vector $\balpha_i$ as follows by $\alpha_{ik} = \exp(- (k - u_i)^2 / \sigma^2) * s, k=0, \ldots, 255$. Then the Dirchlet distribution with $\balpha_i$ can gives a probability vector with probabilities concentrated around $u_i$. Vectors $(\bv_i)_{i \in V}$ are computed from a two-layer MLP branching out from the second-to-the-last layer of the U-net. To compute $\bL$, we set $\gamma=5.0$. We treat each pixel with a surrounding $3 \times 3$ window as its neighbors. Except for the learning rate as 0.0001, all hyperparameters are kept in line with \citet{lehtinen2018}.

\section{Approximating Other Distributions: Details and Additional Results}

\subsection{Approximating an MRF}

We give more details about the main paper's Table 6, an experimental comparison between a mean-field variational inference baseline for the MRF, and using our NMM to approximate the MRF.
In this experiment, the function $f(\bX; \theta)$ has two separate CNNs to parameterize $\balpha$ and $\bL$. The input to both CNNs is the graph with MRF potentials attached to graph nodes, so $\bX$ includes all MRF potentials. The first layer of the CNN has kernel size $11 \times 11$ and filter size 128, followed by a ReLU nonlinearity. The second layer of the CNN has the kernel size $1 \times 1$, followed by an identity activation. The number of filters in the second layer is set to be 2 for the $\balpha$ and the number of neighbors (49) for the $\bL$. We treat each pixel with a surrounding $7 \times 7$ window as its neighbors. We train our model using Adam optimizer \citep{kingma2014} with 1000 maximum epochs and $0.001$ step size.

\subsection{Approximating Deep Generative Graph Models
with Latent Variables}
\subsubsection{Experimental Details}

We follow the same experiment setting as in DGLFRM \citep{mehta2019}. The only difference is that DGLFRM samples $\by_i$ from the probability vector $\bm{\pi}_i$ directly, while we sample $\by_i$ from $\bm{\pi}_{c_i}$ with $c_i \sim \mathrm{Categorical}(\bL_i)$ to describe neighbor correlations. To compute $\bL_i$ from $\bv_i$, we use the same amortization structure as DGLFRM. Specifically, the inference network for $(\bv_i)_{i \in V}$ is a two layer GCN. The first layer has 32/64 hidden nodes for Cora, Citeseer and Pubmed, or 128/256 hidden nodes for NIPS12 and Yeast. The second layer has 50/100/200 hidden nodes to represent the community size. All the models are trained for 500 to 1000 iterations. Adam optimizer is used at a learning rate 0.01.

\begin{figure}[!t]
    \centering
    \includegraphics[width=0.4\linewidth]{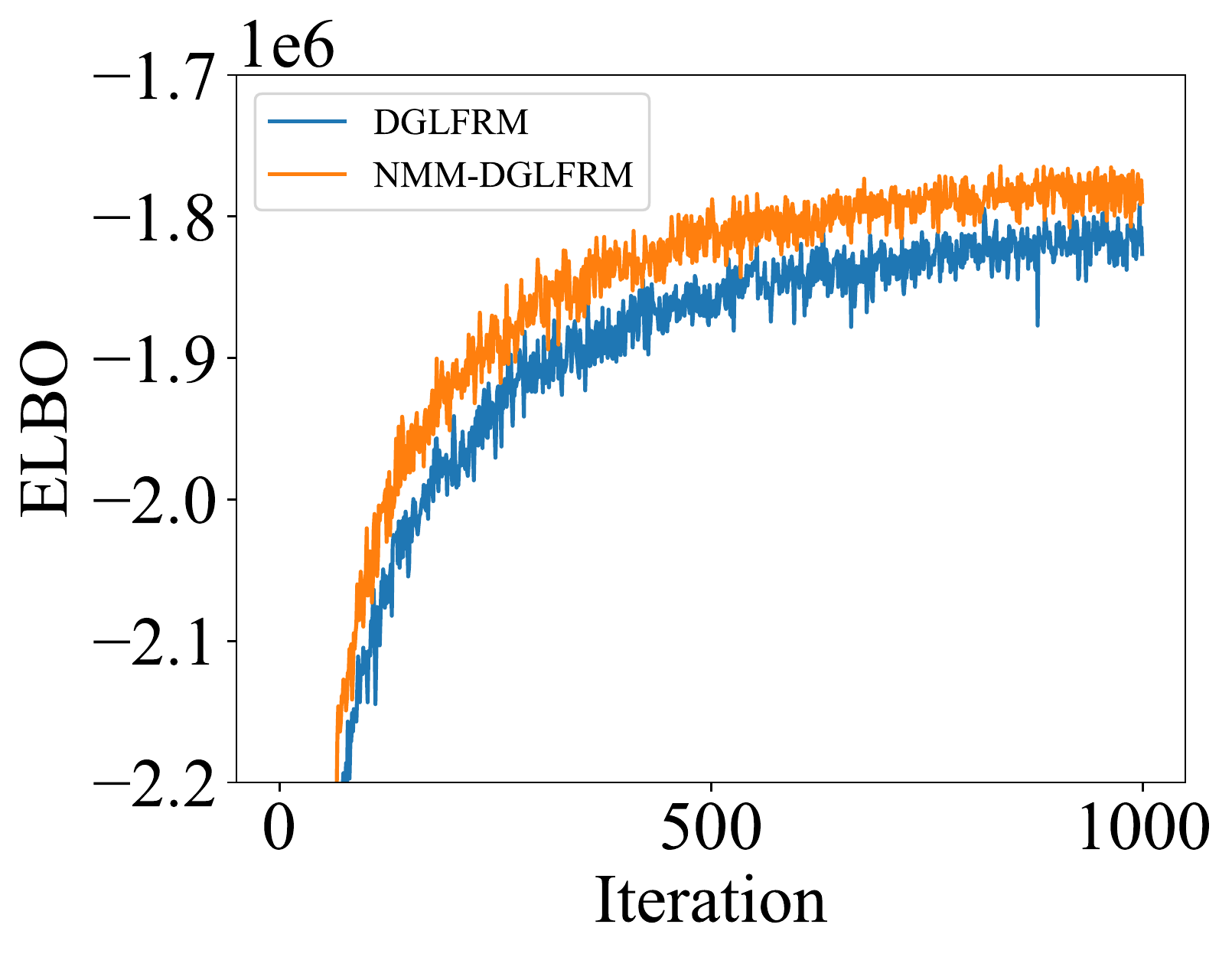}
    \caption{ELBO curve as a function of number of iteration. Tested on NIPS12.}
    \label{exp-elbo}
\end{figure}

\subsubsection{Additional Results: Link Prediction}
\begin{table*}[ht]
\centering
\scalebox{0.88}{\begin{tabular}{ cccccc } 
\hline
\textbf{Algorithm} & \textbf{NIPS12} & \textbf{Yeast} & \textbf{Cora} & \textbf{Citeseer} & \textbf{Pubmed} \\
\hline
DGLFRM & \textbf{86.47 $\pm$ 0.83} & 77.97 $\pm$ 0.79 & 93.00 $\pm$ 0.34 & \textbf{93.79 $\pm$ 0.42} & \textbf{96.11 $\pm$ 0.19} \\
\hline
NMM-DGLFRM & \textbf{86.88 $\pm$ 0.67} & \textbf{79.06 $\pm$ 0.72} & \textbf{93.48 $\pm$ 0.41} & \textbf{93.81 $\pm$ 0.50} & \textbf{96.05 $\pm$ 1.16} \\
\hline
\end{tabular}}
\caption{Results of AUC ROC.}
\label{exp-sbm-roc}
\end{table*}
The main paper's Table 4 showed link prediction results as measured by Average Precision (AP). Additionally, here we share similar link prediction results evaluated by the  Area Under the ROC Curve (AUC). Table \ref{exp-sbm-roc} compares results from the original DGLFRM and results from our inference method. NMM inference significantly improves the performance of DGLFRM on two out of five datasets (Yeast and Cora).

\subsubsection{Additional Results: Likelihood Bound Quality}
As we have mentioned in the submission, using the NMM for inference can better maximize the variational lower bound on the marginal likelihood of observed labels under the DGLFRM model than the original inference method. The trace plot in Figure \ref{exp-elbo} shows evidence of this improvement, showing a noticeably higher value of the evidence lower bound (ELBO). From this, we conclude that sharing neighborhood information with our proposed NMM provides improved modeling capabilities for graph data.

\end{document}